\pdfoutput=1
\documentclass{article}

\usepackage[utf8]{inputenc} % allow utf-8 input
\usepackage[T1]{fontenc}    % use 8-bit T1 fonts
\usepackage{hyperref}       % hyperlinks
\usepackage{url}            % simple URL typesetting
   
\usepackage{amsfonts}
\usepackage{amsmath,amssymb}
\usepackage{mathtools}      % blackboard math symbols

\usepackage{graphicx}
\usepackage{tabularx}
\usepackage{subcaption}

\usepackage[ruled, vlined, linesnumbered]{algorithm2e}

\title{Aspect Term Extraction using Graph-based Semi-Supervised Learning}
   
\author{Gunjan ~Ansari
   \\
  \texttt{gunjanansari@jssaten.ac.in} 
   \and
Chandni ~Saxena \\
 \texttt{cmooncs@gmail.com} 
  \and
Tanvir ~Ahmad\\
\texttt{tahmad2@jmi.ac.in } \\ 
\and
M.N.~Doja \\
\texttt{mdoja@jmi.ac.in} 
\and 
 Department of Computer Engineering,
 \\Jamia Millia Islamia, 
 \\New Delhi-25, India }
\begin{document}
\maketitle

\begin{abstract}
Aspect based Sentiment Analysis is a major subarea of sentiment analysis. Many supervised and unsupervised approaches have been proposed in the past for detecting and analyzing the sentiment of aspect terms. In this paper, a graph-based semi-supervised learning approach for aspect term extraction is proposed. In this approach, every identified token in the review document is classified as aspect or non-aspect term from a small set of labeled tokens using label spreading algorithm. The k-Nearest Neighbor (kNN) for graph sparsification is employed in the proposed approach to make it more time and memory efficient. The proposed work is further extended to determine the polarity of the opinion words associated with the identified aspect terms in review sentence to generate visual aspect-based summary of review documents. The experimental study is conducted on benchmark and crawled datasets of restaurant and laptop domains with varying value of labeled instances. The results depict that the proposed approach could achieve good result in terms of Precision, Recall and Accuracy with limited availability of labeled data.  
\end{abstract}

%\begin{keyword}
%Aspect based sentiment analysis\sep graph-based semi-supervised learning\sep opinion summary\sep label spreading. 

%\end{keyword}
%\cortext[cor1]{Corresponding author. Tel.: +852 6301 2295}
%\end{frontmatter}

%\enlargethispage{-7mm}
\section{Introduction}
\label{section1}
The tremendous increase in the social media content during the past decade has led to rising research trends in the area of Sentiment Analysis (SA). Sentiment Analysis or Opinion mining is gathering, analysis and aggregation of online content available on various blogs, e-commerce sites, movie review sites etc. Nowadays, before buying any product the customer depends heavily on the review content posted by others on various e-commerce sites. The increasing volume and variety of online content makes it difficult for the customer to manually go through the content for decision making. Thus, opinion mining is required for effective customer and organizational decision making. The major task of Sentiment Analysis is to determine the polarity of expressed opinion in the review content. The researchers are developing many tools and techniques to aid in this task.

The growth in the area has resulted in the emergence of various subareas addressing different level of analysis~\cite{b1}. Aspect level sentiment analysis is a subarea of Sentiment Analysis, whose major focus is to determine the sentiments of products/service at aspect level rather than sentence or document level. The attributes of any entity referred in the review is termed as `\emph{aspect}'. Aspect level SA consists of two phases: Aspect term Extraction \cite{b2} and Polarity detection of identified aspect term~\cite{b3}. For example, in the review sentence ``\emph{Battery life of phone is very good but display quality is quite poor}''. The sentence expresses two different opinions on extracted aspect terms `\emph{battery life}' and `\emph{display quality}' of same entity `\emph{phone}'. Sentiment analysis at aspect level will help users to get insight of the product which will enable them to get comparative views of different products while making decision. Also based on the analysis of buyers' views on various aspects, the organizations can improve the quality of their product or service. 
Some of research in the area of Aspect based Sentiment Analysis were summarized in the study~\cite{b4}. In early 2000, works in the area of aspect term extraction were frequency and syntax based approaches. The drawback of frequency based approach is that it identifies many false aspect terms by considering all frequently occurring nouns and noun phrases as aspects. The syntax based approaches can also extract low frequency aspect term by using a simple adjectival modifier relation between opinion word and aspect term. But with growth of data, these approaches fail to perform with accuracy and efficiency. Supervised learning has been employed by the researchers in the recent years due to the availability of few labeled datasets. Supervised learning algorithm has a drawback as it can not learn from unlabeled data. 

In this paper, an approach is proposed that employs a graph-based semi-supervised learning for aspect term classification. In the first step of the proposed approach, a graph $G$ is constructed that consist of set of vertices $V$ which are represented by tokens in a review and weighted edge $E$ between two vertices is represented by the distance between two tokens. For implementation, every token in the review document is represented as a feature vector and the similarity between these tokens is used to represent adjacency matrix of the constructed graph. In the next step, graph sparsification using K-nearest neighbor is employed to make the approach more time and memory efficient. Finally, a label spreading algorithm is used to infer the label of unlabeled tokens in the graph.

The advantage of the the proposed approach is that it does not require human effort for annotating large amount of data thus saving time and expense as contrast to supervised learning approaches. The identification of robust and pruned feature set results in better representation of this model. The work is further extended to determine the polarity of the opinion words associated with the identified aspect terms in the review sentence to generate visual aspect-based summary of review documents. This visual summary presented in this study is quite useful for the customers and manufacturers for knowing the pros and cons of the product or service without going through the whole set of review documents. The experimental study on benchmark and crawled datasets of restaurant and laptop domains depict that the proposed model for aspect term classification could achieve good results in terms of Precision, Recall and Accuracy with only $10\%-20\%$ labeled instances.    
The main contributions of the proposed work are as follows:
\begin{enumerate}%[i]
\item Identification of robust feature set for aspect term extraction.
\item A novel adaptation of  graph-based semi-supervised learning for aspect term detection.
\item Opinion summarization of extracted aspect terms 
\item Performance analysis of the proposed approach with extensive experimental results on four data sets.
\end{enumerate}
  
The rest of the paper is organized as: some of the related works in the area of aspect level sentiment analysis are described in section~\ref{sec2}. Section~\ref{sec3} consists of overview of the proposed work. The experimental setup and performance analysis are shown in section~\ref{sec4} ans section~\ref{sec5} respectively. Finally section~\ref{sec6} concludes the proposed work.
\section{Related work}
\label{sec2}
There are many existing works in the area of Aspect based Sentiment Analysis. The work in Hu et al.~\cite{b5} proposed a feature-based review summarizer for product reviews. The authors identified frequent feature set of maximum three words using association rule mining. Some of the incorrect features that also occur frequently are removed using compactness and redundancy pruning to remove incorrect features from identified feature phrases. They extracted all adjectives near the identified features as opinion word and for each opinion word its semantic orientation is found to determine overall opinion related to the review sentence. Opinion observer proposed in the study~\cite{b6} was a framework designed for comparing two products on similar set of features. The visual summarizer presented in their work could give a clear picture of the strengths and weaknesses of any product to the customer and manufacturer. The pros and cons written by the reviewers are analyzed to extract product features in their technique. Zhuang et al.~\cite{b7} integrated \emph{WordNet}, statistical analysis and movie knowledge to propose a multi-knowledge based approach for generating feature class-opinion summarization of movie reviews. An effective and flexible approach that utilized syntactic and semantic information to extract product feature and their opinion was proposed in their work~\cite{b8}. The authors used dependency tree to extract the relationship between head or governor word and modifier or dependent for the task of feature-opinion mining. An unsupervised rule-based approach for extracting implicit and explicit features from product reviews was presented ~\cite{b9}.  In this work authors exploited common sense knowledge and dependency tree for feature mining.

In the recent years with the increase in the volume and variety of online data, machine learning approaches for aspect term detection and opinion mining have gained popularity. Mukherjee et al. ~\cite{b10} proposed two statistical models named \emph{Seeded Aspect and Sentiment (SAS)} and \emph{Maximum Entropy based SAS (ME-SAS)} for aspect classification. The first method uses some seed words according to user interested category as input before learning to determine aspect terms whereas ME-SAS approach requires no user input before Maximum Entropy training. The authors also employed \emph{Chi-Square} feature selection method for improving the performance of classifier. An unsupervised approach for aspect extraction proposed in ~\cite{b11} learns prior knowledge from big data of reviews available on the web. Authors applied Latent Dirichlet Allocation based topic modeling for extraction of coherent aspect terms. The model can learn from diverse domains and is fault tolerant. Another lifelong machine learning work ~\cite{b12} utilized relaxation graph labeling algorithm to perform unsupervised belief propagation on graph for aspect term extraction. The novelty in this work is that authors utilized past knowledge learned for classification of target as aspect, entity or NIL in the current domain.

The supervised lifelong learning approach for aspect term extraction using \emph{Conditional Random field} is proposed in ~\cite{b13}. A supervised learning approach for determining the aspect terms in a sentence and their polarity is applied~\cite{b14} on \emph{SemEval2014 ABSE} task. The authors proposed CRF model for aspect term extraction and Multinomial Na\"ive Bayes classifier for their polarity detection. On the same dataset, another supervised method using Na\"ive Bayes Classifier for aspect term and its polarity detection is proposed in ~\cite{b15}. The graph-based approaches have been employed by the researcher~\cite{b16} in the past for the task of sentiment categorization. This approach in this paper proposes a novel adaptation of this learning approach for aspect term classification as discussed in the next section. 
\section{Proposed work}
\label{sec3}
\begin{figure}
\centering
\includegraphics[scale=0.20]{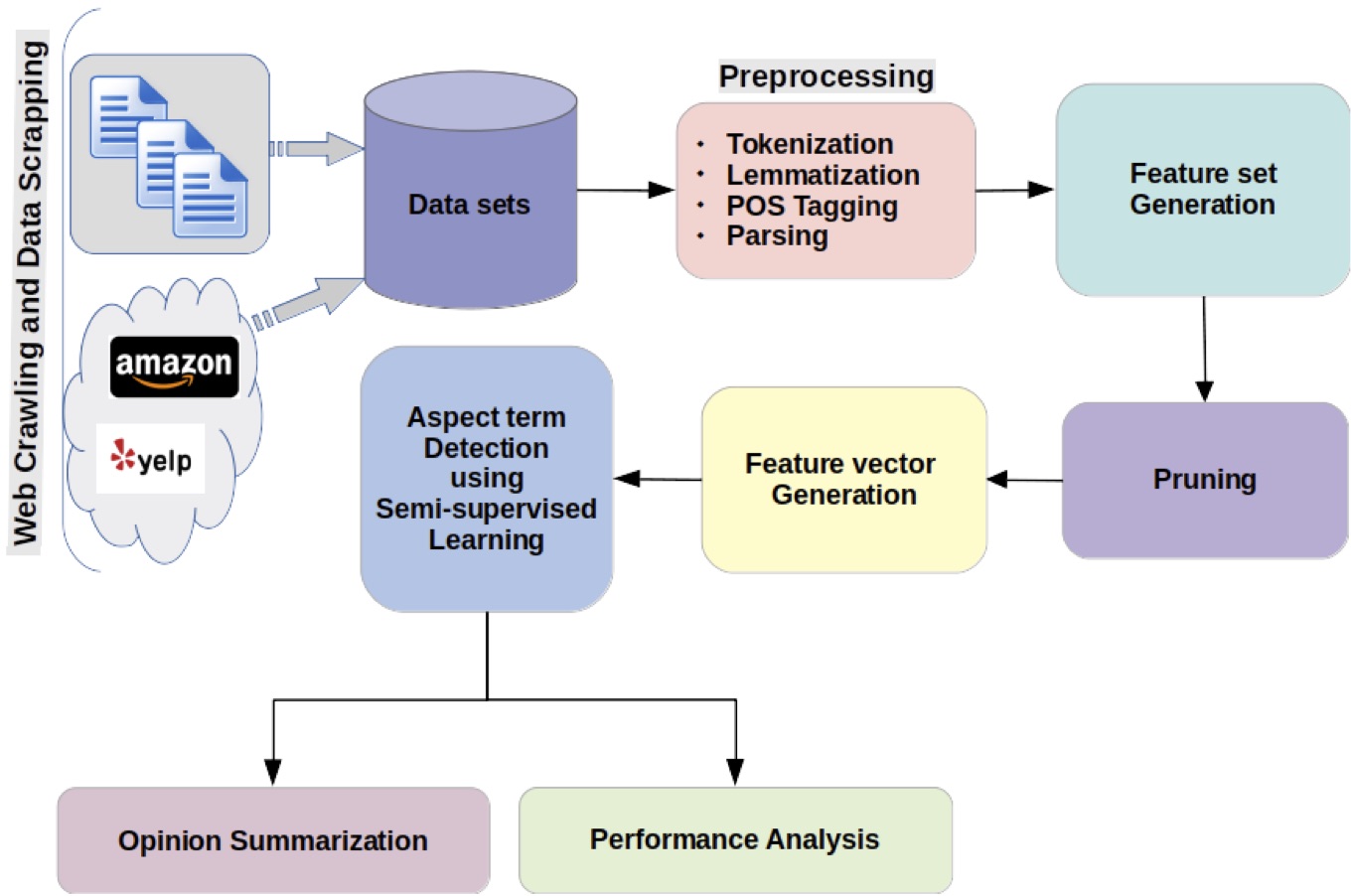}
\caption{Architecture of proposed method}
\label{fig1}
\end{figure}

The architectural view of the proposed model is shown in fig.\ref{fig1}. The model performs aspect term extraction from few tokens labeled into aspect and non-aspect class. The aspect term classification is performed using \emph{k-nearest neighbor (kNN)} based semi-supervised learning. The sentiment class of the generated list of aspect terms is detected by analyzing the polarity of the opinion words associated with them. The details of the proposed model is described in the following subsections.
\subsection{Data collection}
In the proposed approach, two types of dataset are used: labeled and unlabeled. The labeled dataset consists of laptop and restaurant. For each review sentence of this labeled set, the aspect terms and their polarity are given. The unlabeled dataset is crawled from the review sites: \emph{amazon.com} and \emph{yelp.com}. The data is preprocessed to remove \emph{XML-tags} to extract relevant information. The relevant information is stored in dataset and analyzed further for generating feature set for aspect term classification.
\subsection{Data Preprocessing }
The review document is sentence tokenized and on each sentence, \emph{Parts-of-Speech (POS)} tagging is performed to assign tags such as \emph{NN} for Noun, \emph{ADJ} for Adjective,\emph{ ADV} for Adverb etc to every token in a sentence. The dependency relation of the words in the sentence is extracted from the dependency parser tree.
\subsection{Feature set Generation}
In this section, specific features used in the proposed work for aspect term extraction are discussed. After rigorous analysis of the datasets of both restaurant and laptop domains, some rules were identified to define robust and common features set for representing every token in a review sentence as a feature vector for aspect term classification. The following features set used in this study have been proved effective for many natural language processing problems.
\subsubsection{Word length}  The length of the token is an important attribute in determining an aspect term. The token whose length is greater than $ 3 $ is assigned $ 1 $ in the feature value of word length.
\subsubsection{Parts-of-Speech tagging} It was found by many researchers~\cite{b5,b6} that majority of aspect terms are tagged as noun. This is a significant parameter for detecting aspect terms. Value  $ 1 $ is assigned to all tokens whose tag is `\emph{Noun }'in POS tagging of the sentence. 
\subsubsection{Frequent aspect terms}The frequency of the aspect term plays a significant role as aspect terms are more frequent than other terms and less frequent than stop words or common words. Thus, two cut off values were decided from training data and a value $1$ is assigned to all those tokens whose frequency lies between these cut off values. For example, the word \emph{"laptop"} is very frequent in laptop dataset and words \emph{"battery life", "display"}  are within a threshold value. If the token is noun, it is assigned value $ 1 $ in its feature vector.
\subsubsection{Head word}The head words are the last words of the noun phrase and they have high probability of being an aspect terms. The value $ 1 $ is assigned to all tokens which are head word of the noun phrase. For example, the noun phrase ``the Asian salad'' has ``salad'' as a head word which is and aspect term for restaurant domain.
\subsubsection{Orthographic features}Some of the aspect terms are written in Capital letters by reviewers to get more focus. Thus, this feature is set as $ 1 $ for all tokens who start with capital or have all upper-case letters.
\subsubsection{Stop-word}It was found that stop-words can never be the aspect terms, thus a feature is assigned $ 1$ if word is stop word else $ 0 $.
\subsection{ Pruning}This step is performed to prune the repeated and irrelevant tokens. In this step, tokens which have frequency more than a threshold are first removed. The repeated tokens are combined to generate unique token vector for the next step. The multi-word tokens are identified by compound relation of the dependency parser terms and these tokens are combined together in this step. This step reduces the number of tokens to a significant value before feature-vector generation. For example noun tagged token ``battery'' and ``life'' are combined into multi-word token ``battery life'' in this step.
\subsection{Feature-vector generation for all tokens}In this step, feature vector is generated for every token using the identified feature set. The feature vector is a Boolean vector of size $ m \times n $ where m is the number of tokens after pruning and n is the number of selected features. Each feature vector of token is assigned aspect class 0 if it is not an aspect term, $ 1 $ if token is an aspect term or $ -1 $ if its aspect class is unknown. This feature vector is then passed to semi-supervised learner to identify the class of unlabeled tokens from small set of labeled tokens.
\subsection{Aspect term detection using graph-based Semi- Supervised learning} 
The graph-based semi-supervised learning utilized for the proposed solution of the  problem  is described in the following subsections:
\subsubsection{Problem setup} 
Given a set of data points or tokens $X={x_1, x_2\cdots . x_l, x_{l+1}\cdots . x_{l+u}}$ represented by feature vector and a label set $L= {1, 2,\dots . c}$. The  problem of aspect term detection is binary classification problem, have $c=2$. Further,  $Y_L = [y_1, y_2\dots  .y_l]$ are the class label of the labeled tokens and $Y_U=[y_{l+1}, y_{l+2} \cdots. y_{l+u} ]$ be the set of unlabeled tokens for $l\lll u$. Each node $x_i$ is represented by the D-dimensional feature vector. The problem is to infer label of the unlabeled nodes given $X$ and $Y_L$ such that data points that are closer have similar labels.
\subsubsection{Graph construction for semi-supervised learning}           
A graph $G= (V,E)$ for the present study is represented by vertex set $X$ and edges $E$ that are weighted  by the adjacency matrix $A$ of size $n\times n$. The adjacency or affinity matrix $A$  of size $n\ times n $ is defined as follows:
\begin{equation}
A_{ij} = \left\{\begin{matrix}
         exp \left( - \sum_{d=1}^{D} \frac{d(x_i,x_j)^2}{2\sigma^2}\right) &if \; i\neq j   \\ \\
        0 &otherwise
\end{matrix} \right.\label{eq1}   
\end{equation}

Where $d(x_i,x_j)$ is the distance metric used to measure the dissimilarity between two feature vectors of token $ i $ and $ j $ and $\sigma$ is the kernel bandwidth parameter. 
\subsubsection{Graph sparsification using K-nearest neighbors (kNN)} 
To increase the efficiency, a memory friendly sparse matrix is constructed from this full connected graph using $kNN$ kernel. In $kNN$ graph, there is a connection between two nodes $i$ and $j$ only if their distance is k-smallest among all others nodes connected to $i$ in set $n-i$. This graph is converted to an undirected weighted sparse matrix $W$. 
\subsubsection{ Label Inference}
The final step of semi-supervised learning is the task of label inference from the constructed graph $G$ and $Y= [y_1, y_2 \cdots y_l, y_{l+1} \cdots .y_{l+u}]$. To diffuse label on unlabeled tokens using semi-supervised learning algorithms, the iterative procedure of Label Spreading algorithm proposed by Zhou et al.~\cite{b17} is described as follows:
\begin{enumerate}%[i]
  \item Construct the diagonal matrix $D= {d_{ii}}$ from the weight matrix $W$ where $d_{ii}=\sum_i W_{ij}$ for all nodes $i$.
  \item Compute the normalized graph Laplacian $S=D^{-1/2} W D^{-1/2} $.
  \item Iterate $Y^{(t+1)}= \alpha S Y^t + (1-\alpha)Y$ until convergence, where $ \alpha $ is the clamping factor having value between $(0, 1)$.
   \item Let $Y^*$ denote the limit of the sequence $Y^t$ .
  \item Finally, label each unlabeled point $x_i$ as a label $y_i = argmax_{j\leq c} Y^*_{ij}$
\end{enumerate}

In Step 1 , diagonal matrix $D$ is constructed by summing up the edge weights from node $i$ to all other nodes $j$ in matrix $W$. In Step 2, matrix  $ W $ is normalized symmetrically using Laplacian matrix and transformed to matrix $S$. This step leads to fast convergence of the iteration. During each iteration of Step 3, Y receives the information from its neighbors as specified in the first term and retains its initial information as specified in second term. The clamping factor determines the relative amount of information that an instance should adopt where $\alpha = 0$ means keeping the initial label information and $\alpha = 1$ means replacing all initial information. So a value between $0$ and $1$ is chosen for $\alpha$ to balance information of both the terms. After convergence of the algorithm, according to the maximum number of allowed iterations or convergence tolerance value, the final label of unlabeled tokens is stored in $Y^*$ as shown in Step 4 and 5. Finally, the tokens labeled as an aspect term are stored in a list for opinion summarization which is covered in next subsection.

\subsection{Opinion summary of extracted aspect terms} The final task of the proposed model is polarity detection of the identified aspect terms from crawled reviews to generate visual aspect based summary. This module summarizes the opinion posted by reviewers on the aspect terms rather than detecting the polarity of the review sentences. The aspect terms stored in the list are scanned in the review sentences. If the term is found, its related opinion words and the associated modifiers are searched in the review sentence. For example, in the review sentence ``\emph{Battery is very good but screen quality is poor}'', the aspect terms detected by the semi-supervised learner are ``\emph{Battery}'' and ``\emph{screen quality}'' and opinion words and modifiers associated with these aspect terms are $<$ good, very $>$ and $<$ poor, NIL $>$ respectively.  The opinion on aspect term is found to be positive, negative or neutral by using \emph{StanfordCoreNLP} library. It  assigns a score to each opinion word in the range of $0-4$, $0$ for very negative, $1$ for negative, $2$ for neutral, $3$ for positive and $4$ for very positive.  The modifiers if present are used to intensify or diminish the opinion words associated with the aspect terms. The visual opinion summary of frequent aspect terms is generated by aggregating the polarity of these terms in all the review sentences of the dataset.
\section{ Experimental setup} 
\label{sec4}   
The experiments on four different datasets analyze and validate the performance of the proposed model. In this section, an overview of the datasets used, pre-processing tools and performance metric used in the experimental study are presented.   

\subsection{Data sets} 
Two benchmark datasets of SemEval 2014 which is publicly available for research purpose are used in this work.  Also, two real datasets of laptop and restaurant reviews crawled from ``\emph{Amazon.com}'' and ``\emph{Yelp.com}'' respectively are used for the experimental study. The dataset statistics is given in Table 1.
\begin{table}[h]
\caption{Datasets statistics}
\begin{tabular*}{\hsize}{@{\extracolsep{\fill}}llll@{}}
%\hrule
Dataset domains & Sentences  &Aspect words &Non-aspect words\\
%\colrule
Laptop(SemEval data) & 2981 & 608 &3726\\
Restaurant (SemEval data) &789 &464   &1972 \\
Laptop (Amazon data) &323 &126 &845\\
Restaurant (Yelp data) &2601 &436 &4121\\
%\hrule
\end{tabular*}
\end{table}

\subsection{Preprocessing} 
The \emph{SemEval 2014} datasets are prepossessed to remove \emph{XML} tags to extract relevant information from the dataset. The real dataset is crawled from the site using `\emph{bs4}' library of python and its review text is stored in the database. The \emph{Natural Language Toolkit (NLTK)}, a collection of libraries for statistical natural language processing, is used to tokenize the reviews and extract useful information such as lemma, `\emph{Parts-of-Speech}' tags etc. `\emph{StandfordCoreNLP}' has been utilized to extract dependency relation from the sentence. The `\emph{nsubj}' relation is used to find relation between the aspect term and opinion word,`\emph{amod}'relation is used to extract the modifier associated with the opinion word and compound relations are used to extract multi-word tokens from the reviews.

\subsection{ Performance metrics}
Accuracy is the ratio of correctly predicted tuples to the total tuples in the dataset as shown in (\ref{eq4}). Accuracy is a good performance metric only when the dataset have same false positive and false negatives which is not the case in majority of data sets. Therefore, the precision (\ref{eq2}) and recall (\ref{eq3}) metrics  are used  to evaluate the performance of the classifier. Precision is an important metrics and is the ratio of correctly predicted positive or negative tuples to the total predicted positive or negative tuples as shown in (\ref{eq2}). The ratio of correctly predicted positive tuples to the total observations in actual class are computed by Recall as shown in the (\ref{eq3}).
\begin{align}
&{Precision} = {\frac{TP}{(TP+FP)}} \label{eq2}\\
&{Recall} = {\frac{TP}{(TP+FN)}} \label{eq3}\\
&{Accuracy} = {\frac{(TP+TN)}{(TP+FN+TN+FP)}} \label{eq4}
\end{align} 

Here, $TP$ is the positive items that are correctly classified as positive, $TN$ is the negative items correctly classified by as negative, $FP$ is the negative tuple incorrectly classified as positive and $FN$ is the positive items incorrectly classified as negative.
\section{ Performance Evaluation}
\label{sec5} 
To evaluate the performance of proposed graph-based semi-supervised learning approach for aspect term classification, the experimental study was conducted with variation in $k$ nearest neighbors and size of labeled data on all four data sets. Each token $i$ is represented as a feature vector $X_i$ and each $X_i$ is associated with class label $Y_i$ where $Y_i \in {C_1, C_2}$ for labeled tokens and $-1$ for unlabeled tokens, $C_1$ stands for Aspect class and $C_2$ stands for Non-aspect Class. Since semi-supervised learning algorithms learn from small amount of labeled data, therefore ratio of labeled set is varied from $10-20\%$ and the value of $k$ is varied from $1$ to $20$ for all experiments. There is a problem of data skewness in the data sets as the number of tokens belonging to $C_1$ class are much smaller in ratio as compared to tokens belonging to $C_2$ class. Thus, before analyzing the performance of the proposed model, it randomly sample the tokens from $C_2$ class to overcome the drawback of data skewness. The comparative analysis between benchmark and crawled dataset of restaurant and laptop reviews in terms of accuracy, precision and recall for varying value of $k$ and labeled data size $10\%$, $15\%$ and $20\%$ is shown in subsection \ref{sec5.2}. To ensure robustness of the proposed model, several runs were performed and the average performance of the learner is reported as the final result. 

\subsection{Parameter setting} 

In the proposed approach, the following parameters are set for experimental study:
\\ 
The  proposed model use \emph{Kernel= kNN}, number of nearest neighbors  $k = 1-20$, clamping factor $= 0.2$, maximum iterations $ = 700$ and convergence tolerance $ = 0.0001$.

\subsection{Result analysis on datasets}\label{sec5.2}
\begin{figure}[!ht]

            \begin{subfigure}[b]{0.3\textwidth}
                \includegraphics[width=1\linewidth, height=4cm]{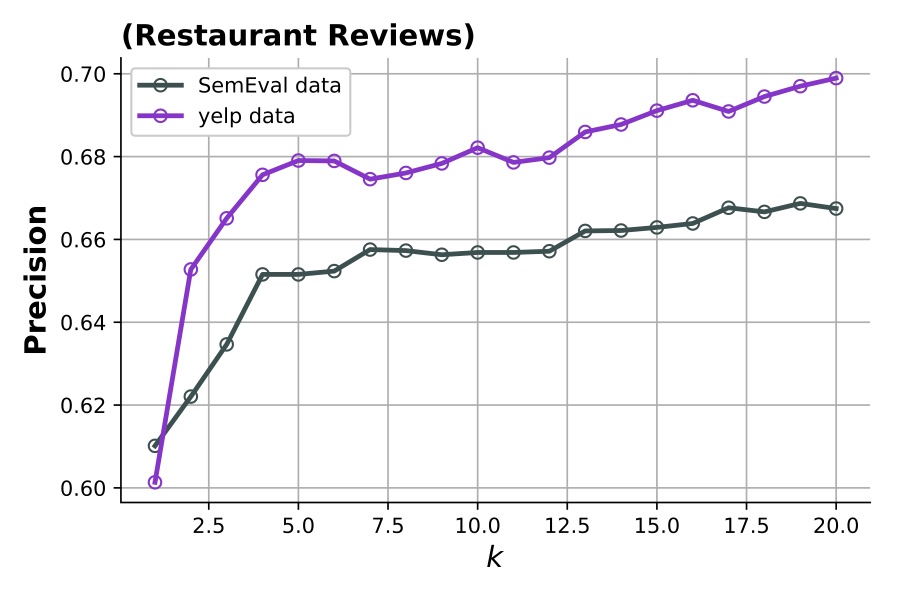}
                \caption{}
            \end{subfigure}
            \begin{subfigure}[b]{0.3\textwidth}
                \includegraphics[width=1\linewidth, height=4cm]{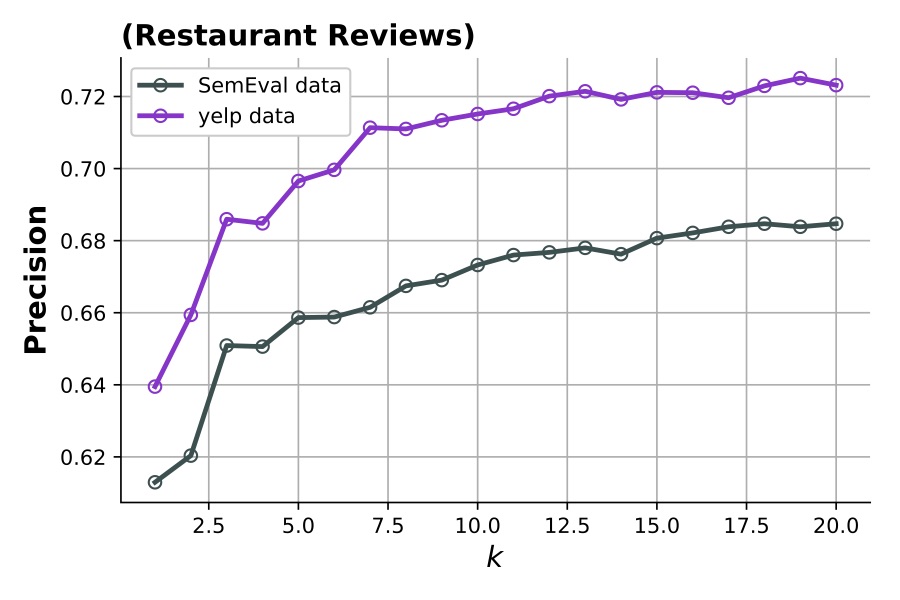}
                \caption{}
            \end{subfigure}
            \begin{subfigure}[b]{0.3\textwidth}
                \includegraphics[width=1\linewidth, height=4cm]{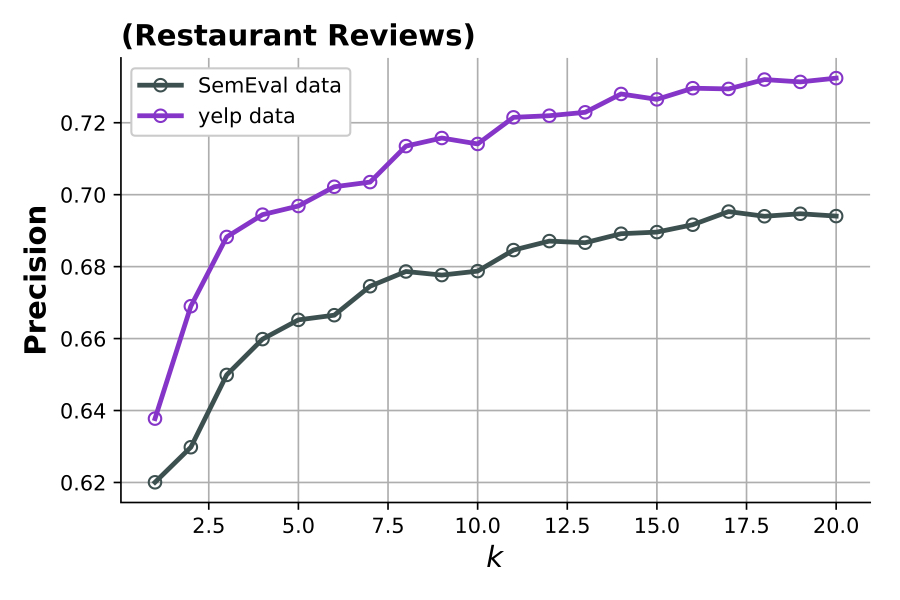}
                \caption{}
            \end{subfigure}
\caption{Performance analysis on Precision metric at (a)10\%, (b)15\% and (c) 20\% labeled data restaurant reviews}  
\label{fig2}           
\end{figure}

\begin{figure}[!ht]

            \begin{subfigure}[b]{0.3\textwidth}
                \includegraphics[width=1\linewidth, height=4cm]{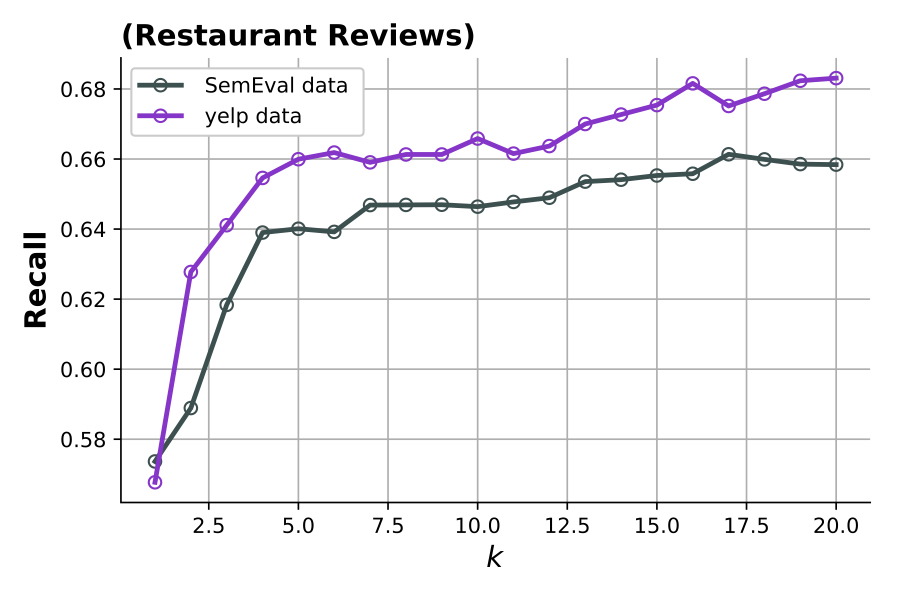}
                \caption{}
            \end{subfigure}
            \begin{subfigure}[b]{0.3\textwidth}
                \includegraphics[width=1\linewidth, height=4cm]{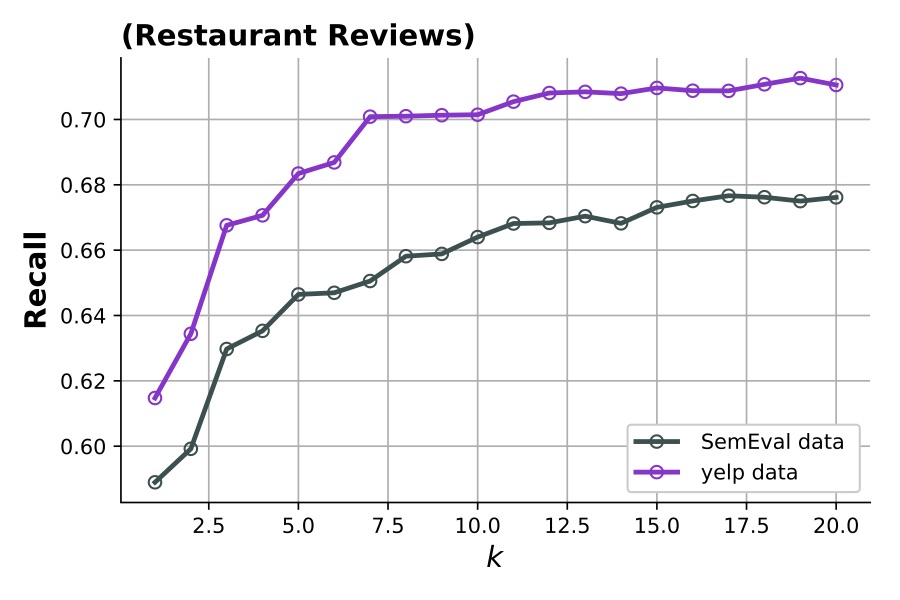}
                \caption{}
            \end{subfigure}
            \begin{subfigure}[b]{0.3\textwidth}
                \includegraphics[width=1\linewidth, height=4cm]{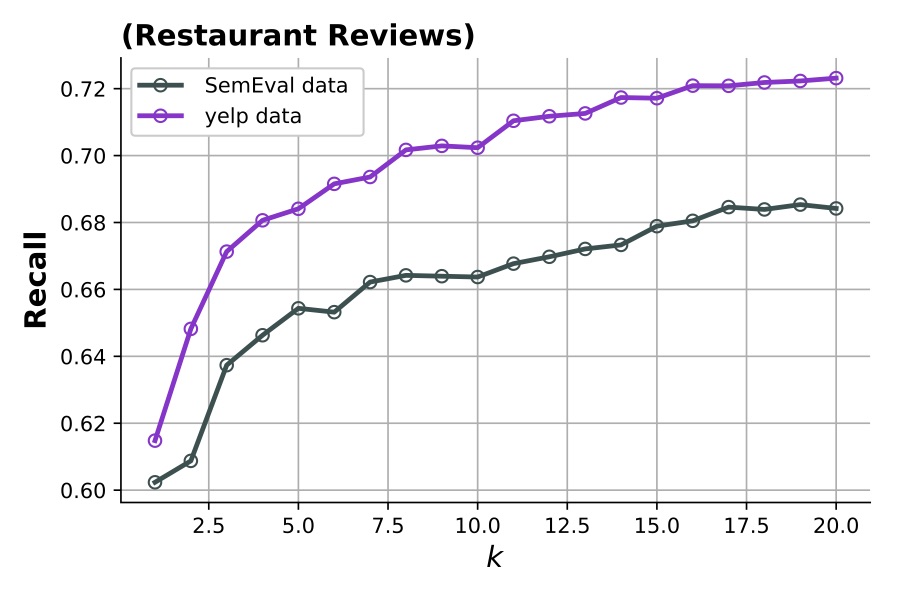}
                \caption{}
            \end{subfigure}
\caption{Performance analysis on Recall metric at (a)10\%, (b)15\% and (c) 20\% labeled data restaurant reviews}  
\label{fig3}           
\end{figure}

\begin{figure}[!ht]

            \begin{subfigure}[b]{0.3\textwidth}
                \includegraphics[width=1\linewidth, height=4cm]{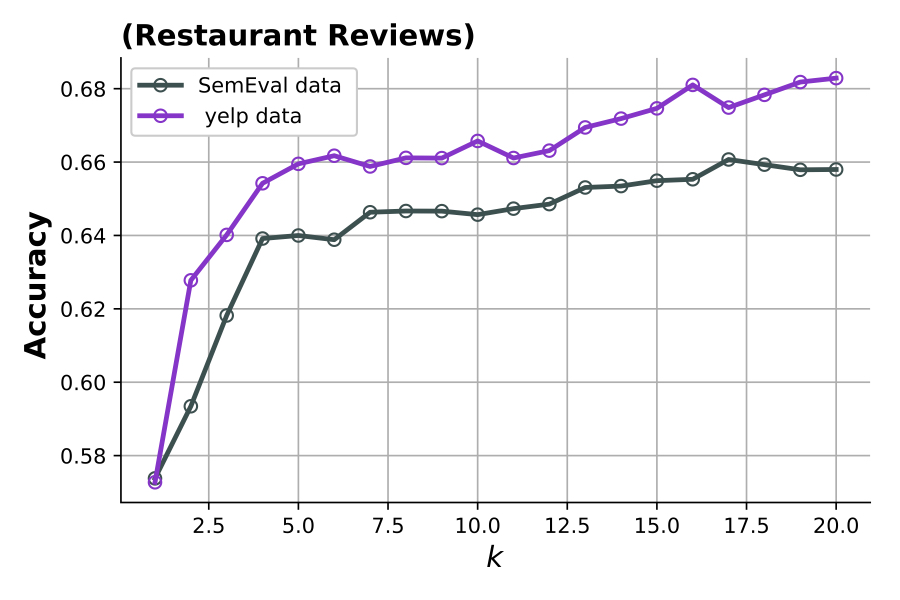}
                \caption{}
            \end{subfigure}
            \begin{subfigure}[b]{0.3\textwidth}
                \includegraphics[width=1\linewidth, height=4cm]{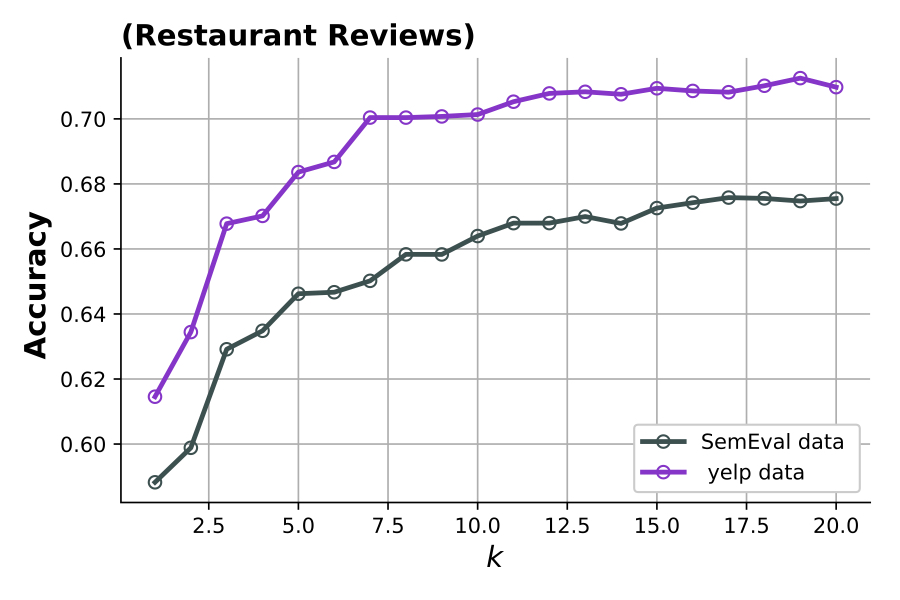}
                \caption{}
            \end{subfigure}
            \begin{subfigure}[b]{0.3\textwidth}
                \includegraphics[width=1\linewidth, height=4cm]{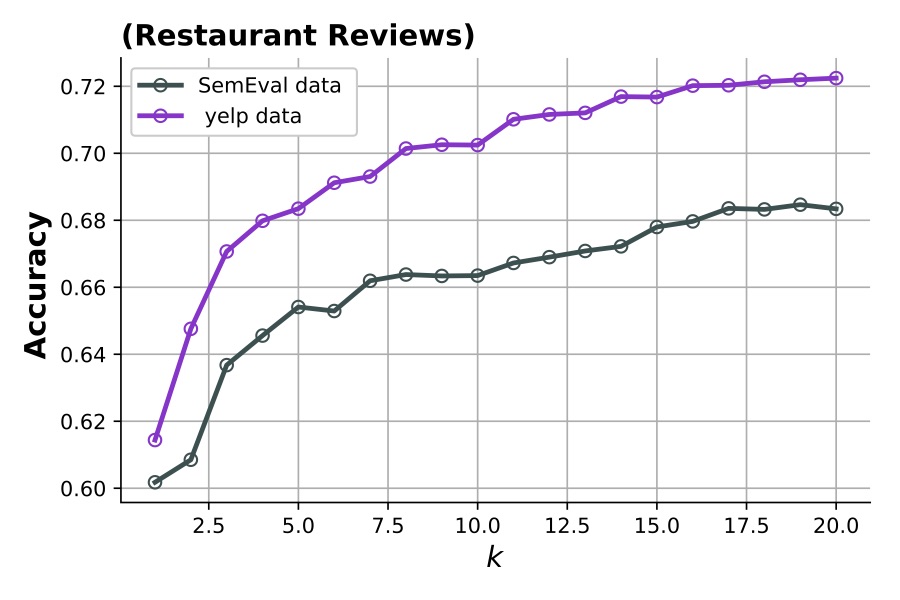}
                \caption{}
            \end{subfigure}
\caption{Performance analysis on Accuracy metric at (a)10\%, (b)15\% and (c) 20\% labeled data restaurant reviews}  
\label{fig4}           
\end{figure}

\begin{figure}[!ht]

            \begin{subfigure}[b]{0.3\textwidth}
                \includegraphics[width=1\linewidth, height=4cm]{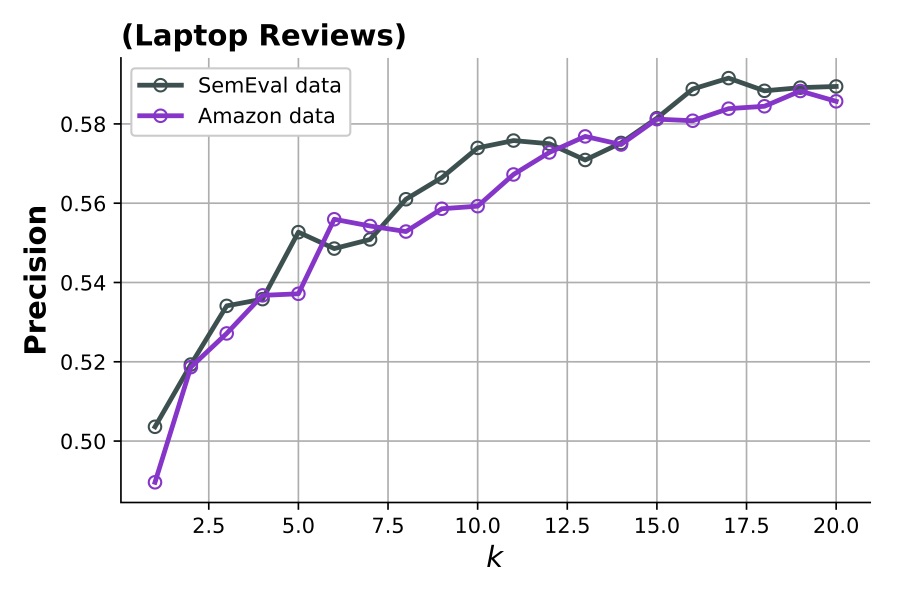}
                \caption{}
            \end{subfigure}
            \begin{subfigure}[b]{0.3\textwidth}
                \includegraphics[width=1\linewidth, height=4cm]{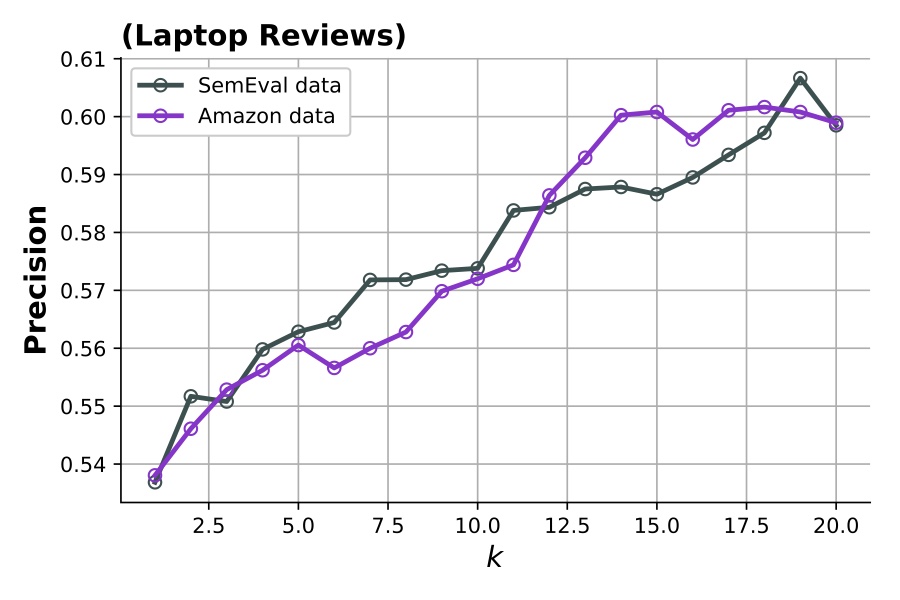}
                \caption{}
            \end{subfigure}
            \begin{subfigure}[b]{0.3\textwidth}
                \includegraphics[width=1\linewidth, height=4cm]{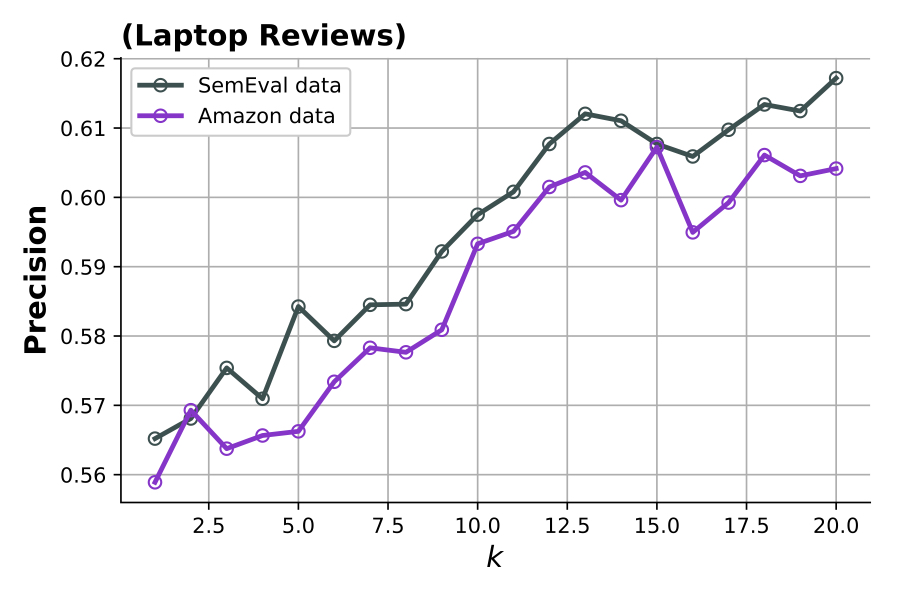}
                \caption{}
            \end{subfigure}
\caption{Performance analysis on Precision metric at (a)10\%, (b)15\% and (c) 20\% labeled data laptop reviews}  
\label{fig5}           
\end{figure}

\begin{figure}[!ht]

            \begin{subfigure}[b]{0.3\textwidth}
                \includegraphics[width=1\linewidth, height=4cm]{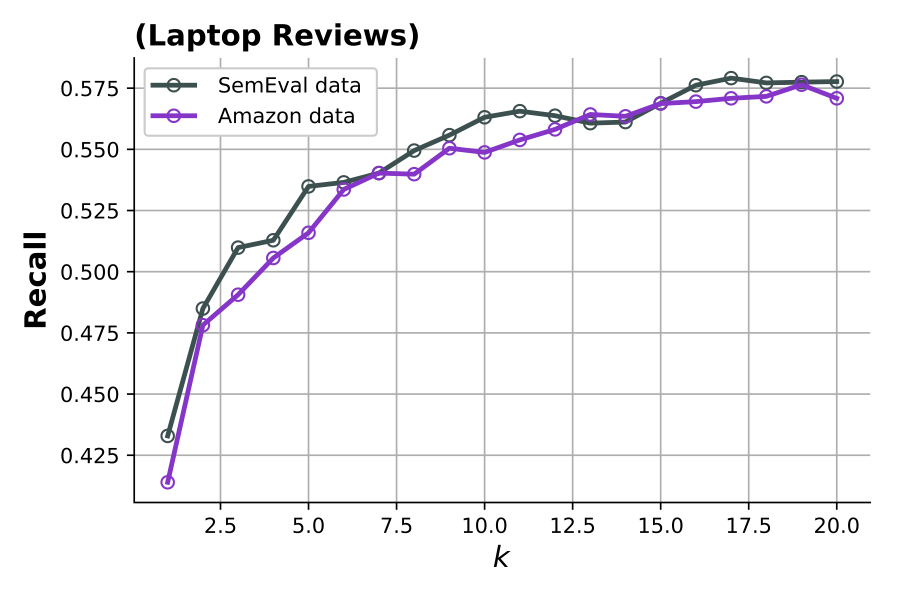}
                \caption{}
            \end{subfigure}
            \begin{subfigure}[b]{0.3\textwidth}
                \includegraphics[width=1\linewidth, height=4cm]{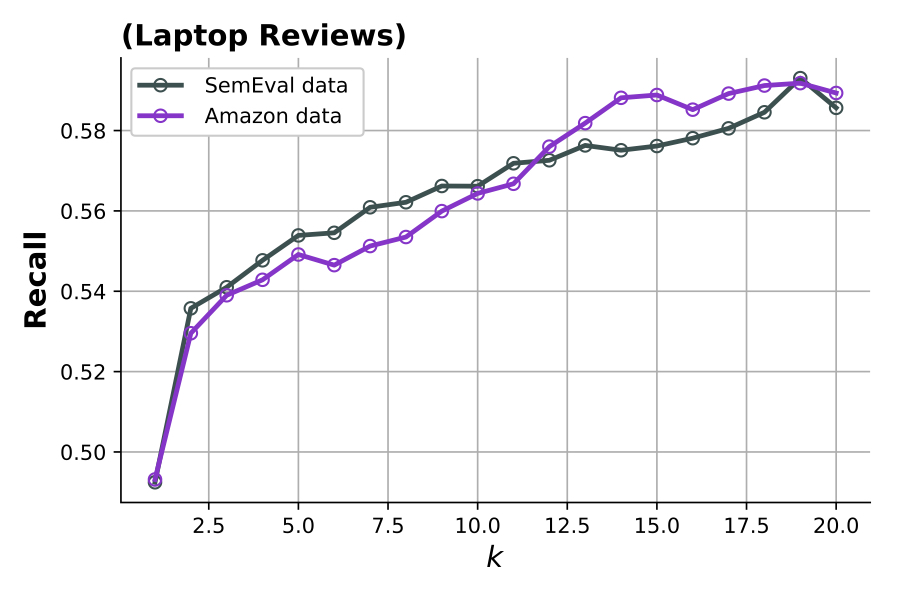}
                \caption{}
            \end{subfigure}
            \begin{subfigure}[b]{0.3\textwidth}
                \includegraphics[width=1\linewidth, height=4cm]{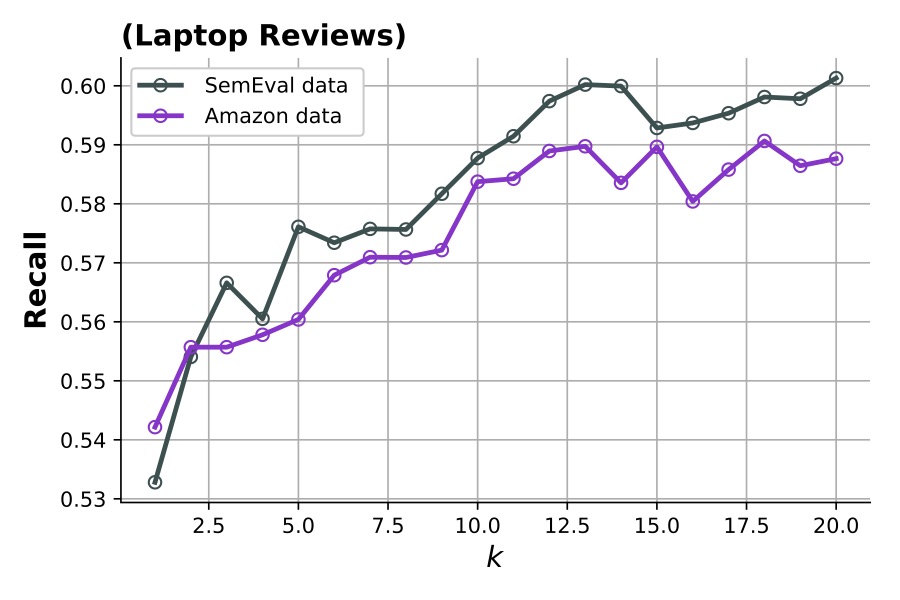}
                \caption{}
            \end{subfigure}
\caption{Performance analysis on Recall metric at (a)10\%, (b)15\% and (c) 20\% labeled data laptop reviews}  
\label{fig6}           
\end{figure}

\begin{figure}[!ht]

            \begin{subfigure}[b]{0.3\textwidth}
                \includegraphics[width=1\linewidth, height=4cm]{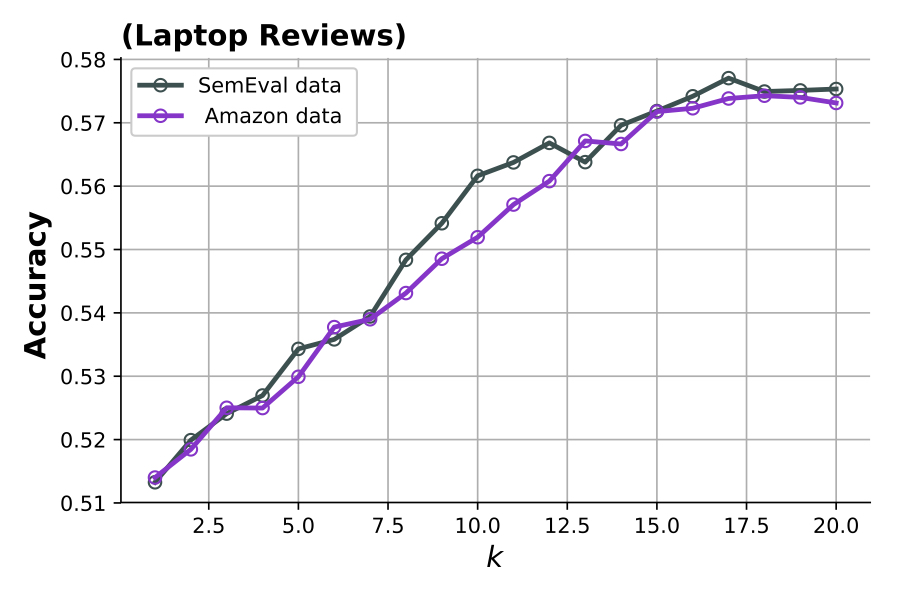}
                \caption{}
            \end{subfigure}
            \begin{subfigure}[b]{0.3\textwidth}
                \includegraphics[width=1\linewidth, height=4cm]{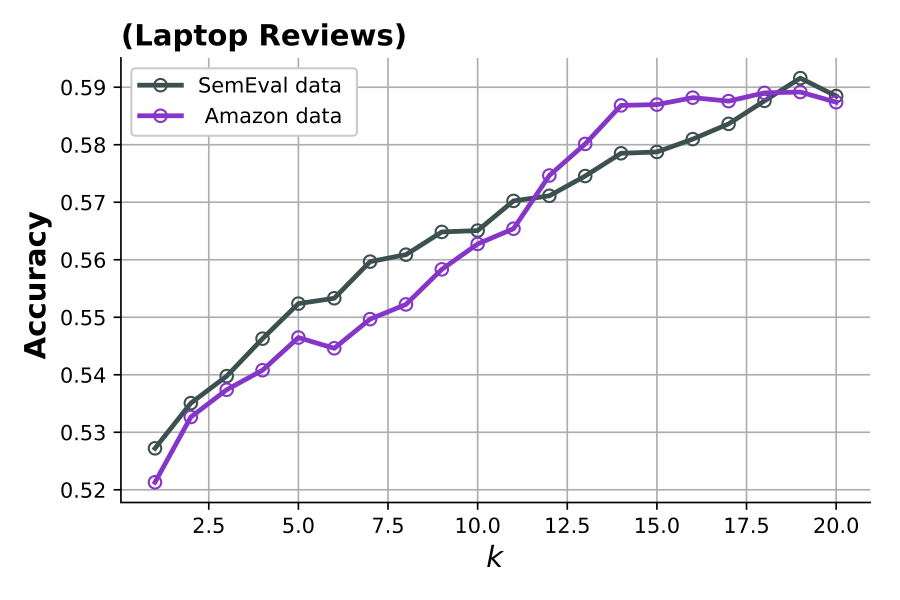}
                \caption{}
            \end{subfigure}
            \begin{subfigure}[b]{0.3\textwidth}
                \includegraphics[width=1\linewidth, height=4cm]{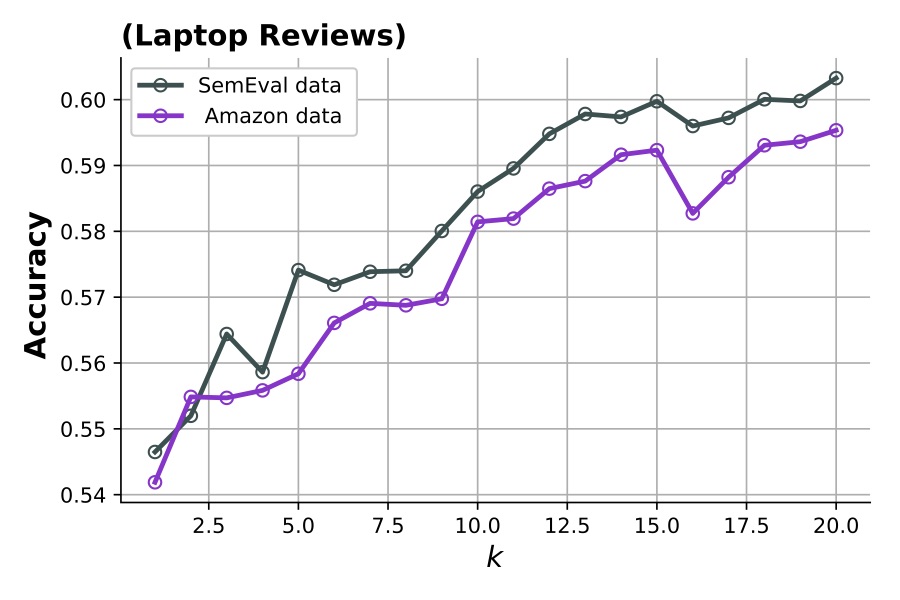}
                \caption{}
            \end{subfigure}
\caption{Performance analysis on Accuracy metric at (a)10\%, (b)15\% and (c) 20\% labeled data on laptop reviews}  
\label{fig7}           
\end{figure}

The comparison between \emph{SemEval} and \emph{Yelp} restaurant reviews in terms of Precision at different values of k and labeled data size can be observed from fig.~\ref{fig2}. The plots in fig.~\ref{fig2} depict maximum precision $\approx73\%$ and $\approx 70 \%$ on Yelp and \emph{SemEval} dataset respectively for labeled dataset size of $20\%$. As observed from plots of fig.~\ref{fig3}, recall value $\approx73\%$ is maximum on Yelp data at $k= 19$. As depicted from plots in fig.~\ref{fig4}, there is an improvement of around $4 \%$ in terms of Accuracy when size of labeled dataset is increased from $10\%$ to $20\%$. In addition, the performance of Yelp dataset is better than\emph{ SemEval} dataset. The reason behind this variation in terms of accuracy is due to less number of non-aspect terms in \emph{SemEval} dataset. Thus, during random sampling the learner had less choice to sample instances of non-aspect class.
fig.~\ref{fig5} shows the comparison between \emph{SemEval} and Amazon laptop reviews in terms of Precision at different values of $k$ and labeled data size. The plots in fig.~\ref{fig5}, \ref{fig6} and \ref{fig7} depict that there is slight improvement of around $2 \%$ in terms of Precision, Recall and Accuracy value when labeled dataset size increases from $10$ to $20\%$. As observed from plots of fig.~\ref{fig7}, maximum accuracy $\approx62\%$ and $\approx60\%$ is reported for\emph{ SemEval} and Amazon reviews respectively.  In this case, learner performed slightly better on \emph{SemEval }dataset than Amazon dataset. The reason behind this drop in accuracy can be observed due to small size of amazon review dataset. 

\begin{figure}[!ht]

            \begin{subfigure}[b]{0.49\textwidth}
                \includegraphics[width=1\linewidth, height=5cm]{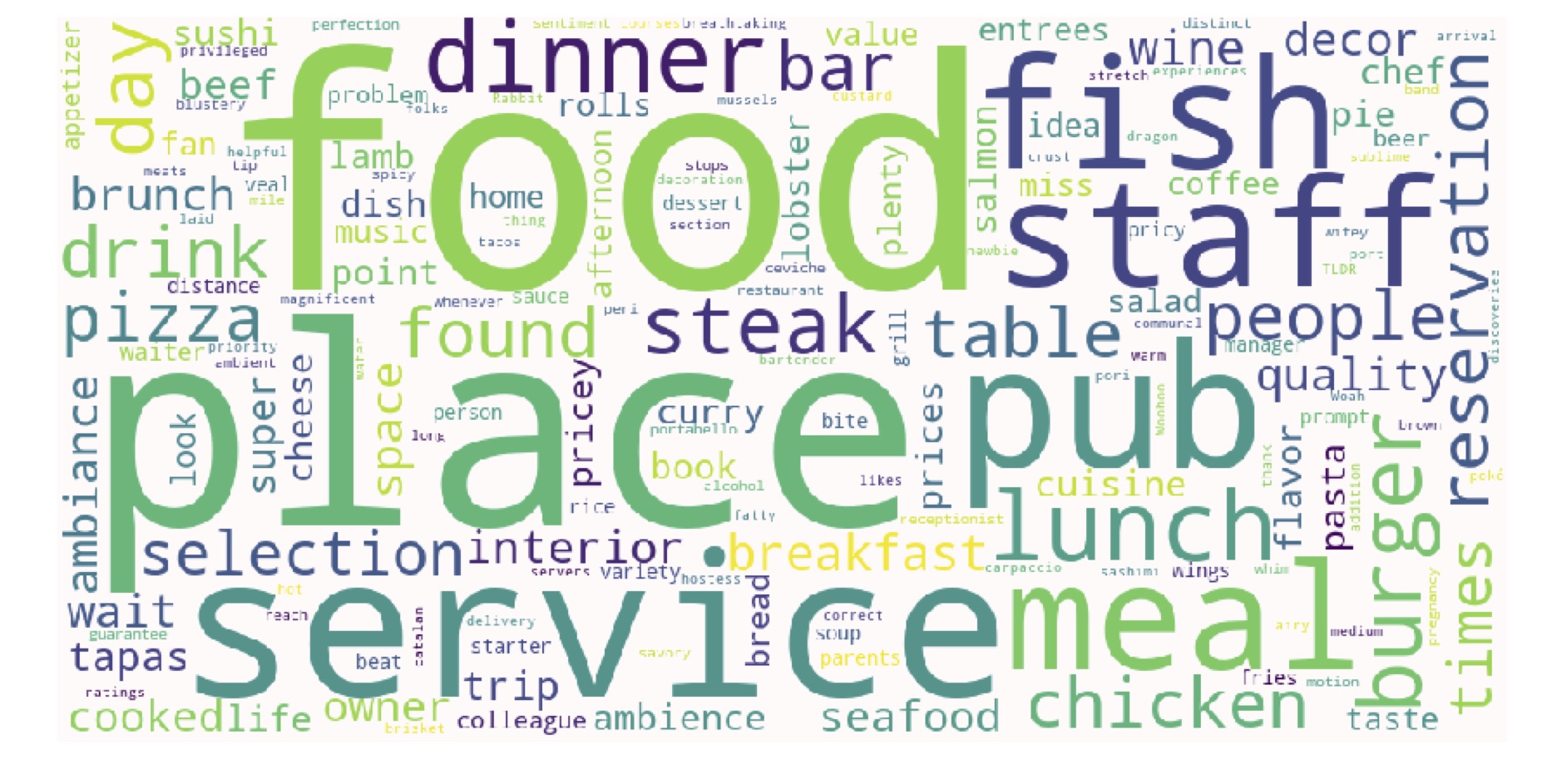}
                \caption{Frequency of aspect terms (yelp data)}
            \end{subfigure}
            \begin{subfigure}[b]{0.49\textwidth}
                \includegraphics[width=1\linewidth]{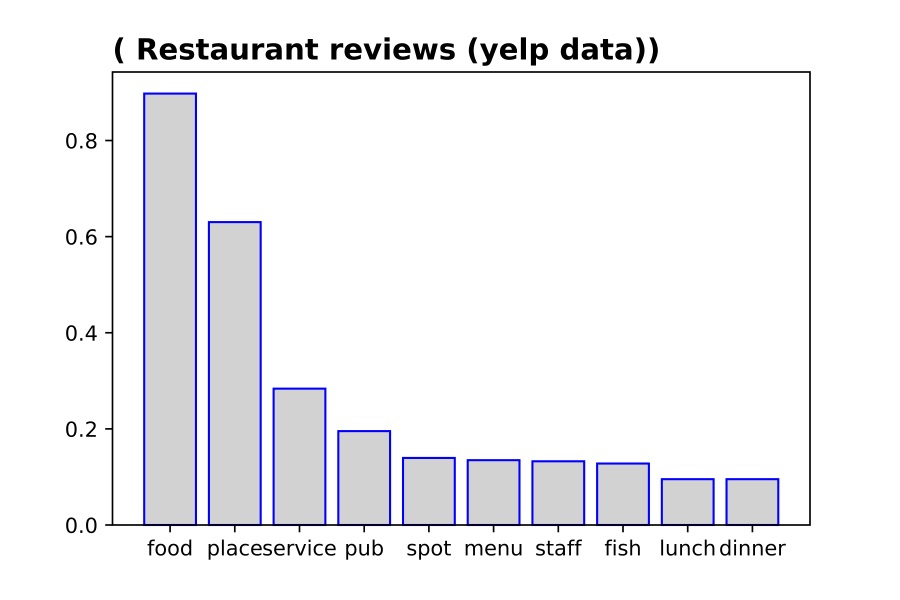}
                \caption{Opinion summary of aspect terms (yelp data)}
            \end{subfigure}
\caption{Aspect terms and their opinion summary (yelp data)}
\label{fig8}
\end{figure}

The aspect terms detected by the proposed model are presented as \emph{Wordcloud} in fig.~\ref{fig8}(a). This visual representation helps the manufacturers or customers in knowing the most popular aspects of the particular product or service. As observed from the fig.~\ref{fig8}(a), `food', `places' and `service' are most frequently occurring aspect terms in the set of review documents. The aspect based summary of the restaurant review dataset crawled from yelp is generated in the final step of the proposed work. The visual opinion summary of ten frequent aspect terms extracted from yelp dataset is shown in fig.~\ref{fig8}(b). The plot depicts that the aspect terms `food',  `places'  and `service' are most liked attributes of the restaurant. As observed from the plot in fig. 9, ‘lunch’ and ‘dinner’ are least liked attributes by most of the reviewers. The attributes `lunch'  and `dinner' getting lowest score despite of `food'  getting highest score is due to the low frequency of these attributes in the reviews and the opinion posted in few reviews on these attributes was negative as well. This opinion summary is quite useful for comparing various products or services in terms of their attributes.


        
\section{Conclusion}
In this paper, a novel adaptation of graph-based semi-supervised learning for aspect term classification is proposed. The time and memory efficient graph representation using kNN makes the approach more suitable for large data size problems. The experimental results on two different domains show that the model achieves good performance with availability of only few labeled instances. The aspect-based summary presented in the proposed work could be beneficial for manufactures and buyers for analysis of different products or services without the need for scanning all reviews in the dataset.

There are several directions to improve the proposed work. In future, the proposed model could be extended to cross-domain aspect term extraction so that the aspect terms learned from one domain can be used to classify the aspect terms of some other domain. This can be performed by identifying domain independent feature set for token representation. The proposed transductive learning approach can further be improved to inductive learning approach by employing Transductive Support Vector Machine (TSVM), an extension of support vector machines with unlabeled data. Bi-partite graph is cost-effective and scalable graph representation that can be used in future to solve semi-supervised learning problem for large and scalable review dataset. 
\label{sec6}


\bibliography{m4.bib}
\bibliographystyle{plain}

\end{document}